\documentclass[a4paper,twoside]{article}
\usepackage[belowskip=-15pt,aboveskip=0pt]{caption}

\usepackage{adjustbox}

\usepackage{longtable}
\usepackage{amsfonts,amsmath,amssymb}
\usepackage{url}
\usepackage{hyperref}
\hypersetup{colorlinks=false,pdfborder={0 0 0}}
\usepackage{colortbl}

\usepackage{epsfig}
\usepackage{subfigure}
\usepackage{calc}
\usepackage{amssymb}
\usepackage{amstext}
\usepackage{amsmath}
\usepackage{amsthm}
\usepackage{multicol}
\usepackage{pslatex}
\usepackage{apalike}
\usepackage{graphicx}
\usepackage{SCITEPRESS}     % Please add other packages that you may need BEFORE the SCITEPRESS.sty package.

\begin{document}
\setlength{\parskip}{0em}
\setlength{\parindent}{0em}

\title{House price estimation from visual and textual features}

\author{\authorname{Eman H. Ahmed, Mohamed N. Moustafa}
\affiliation{Computer Science and Engineering Department, The American University in Cairo, Road 90, New Cairo, Cairo, Egypt}
\email{eman.hamed@aucegypt.edu, m.moustafa@aucegypt.edu}}


\abstract{Most existing automatic house price estimation systems rely only on some textual data like its neighborhood area and the number of rooms. The final price is estimated by a human agent who visits the house and assesses it visually. In this paper, we propose extracting visual features from house photographs and combining them with the house's textual information. The combined features are fed to a fully connected multilayer Neural Network (NN) that estimates the house price as its single output. To train and evaluate our network, we have collected the first houses dataset (to our knowledge) that combines both images and textual attributes. The dataset is composed of 535 sample houses from the state of California, USA. Our experiments showed that adding the visual features increased the R-value by a factor of 3 and decreased the Mean Square Error (MSE) by one order of magnitude compared with textual-only features. Additionally, when trained on the textual-only features housing dataset \cite{Lichman:2013}, our proposed NN still outperformed the existing model published results \cite{BinKhamis_2014}.}

\onecolumn \maketitle \normalsize \vfill

\section{\uppercase{Introduction}}
\label{sec:introduction}

Housing market plays a significant role in shaping the economy. Housing renovation and construction boost the economy by increasing the house sales rate, employment and expenditures. It also affects the demand for other relevant industries such as the construction supplies and the household durables \cite{li2011forecasting}. The value of the asset portfolio for households whose house is their largest single asset is highly affected by the oscillation of the house prices. Recent studies show that the house market affects the financial institutions profitability which in turn affects the surrounding financial system. Moreover, the housing sector acts as a vital indicator of both the economy's real sector and the assets prices which help forecast inflation and output \cite{li2011forecasting}. The traditional tedious price prediction process is based on the sales price comparison and the cost which is unreliable and lacks an accepted standard and a certification process \cite{BinKhamis_2014}. Therefore, a precise automatic prediction for the houses' prices is needed to help policy makers to better design policies and control inflation and also help individuals for wise investment plans \cite{li2011forecasting}.
Predicting the houses' prices is a very difficult task due to the illiquidity and heterogeneity in both the physical and the geographical perspectives of the houses market. Also, there is a subtle interaction between the house price and some other macroeconomic factors that makes the process of prediction very complicated. Some previous studies were conducted to search the most important factors that affect the houses' price. All the previous work was directed towards the textual attributes of the houses \cite{BinKhamis_2014,ng2015machine,Park20152928}. So, we decided to combine both visual and textual attributes to be used in the price estimation process. According to \cite{Limsombunc_2004}, the house price gets affected by some factors like its neighbourhood, area, the number of bedrooms and bathrooms. The more bedrooms and bathrooms the house has, and the higher its price. Therefore, we depended on these factors besides the images of the house to estimate the price. The contribution of this paper:
\begin{itemize}
\item We provide the first houses dataset, to our knowledge, that combines both visual and textual attributes to be used for price estimation. The dataset will be publicly available for research purposes.
\item We propose a multilayer neural network for house price estimation from visual and textual features. We report the results of this proposed model using the newly created benchmark dataset. Additionally, we show that our model surpasses the state of the art models, when tested  using only the textual features, on an existing benchmark housing dataset \cite{Lichman:2013}. Our model also outperforms Support Sector Regression machine (SVR) when trained and tested on our dataset.
\end{itemize}
The remaining of this paper is organized as follows: we start by reviewing related work, followed by a description of our newly created dataset. We then present our proposed baseline NN model. The experimental results section demonstrates the accuracy of our proposed model. Finally, we close with some concluding remarks.

\section{\uppercase{Related work}}

During the last decade, some work has been done to automate the real estate price evaluation process. The successes were in emphasizing the attributes of the property such as the property site, property quality, environment and location. Comparing different methods, we found that the previous approaches can be classified into two main categories: Data disaggregation based models and Data aggregation based models. The Data disaggregation based models try to predict the house's price with respect to each attribute alone like the Hedonic Price Theory. However, The Data aggregation models depend on all the house's attributes to estimate its price such as the Neural Network and regression models.
As an example of the Data disaggregation models, the Hedonic Price Theory where the price of the real estate is a function of its attributes. The attributes associated with the real estate define a set of implicit prices. The marginal implicit values of the attributes are obtained by differentiating the hedonic price function with respect to each attribute \cite{Limsombunc_2004}. The problem with this method is that it does not consider the differences between different properties in the same geographical area. That's why it is considered unrealistic. Flitcher et al in \cite{fletcher2000modelling} tried to explore the best way to estimate the property price comparing the results of aggregation and disaggregation of data. They found that the results of aggregation are more accurate. They also found that the hedonic price of some coefficients for some attributes are not stable, as they change according to location, age and property type. Therefore, they realized that the Hedonic analysis can be effective while analysing these changes but not for estimating the price based on each attribute alone. Additionally, they discovered that the geographical location of the property plays an important role in influencing the price of the property.
For the Data aggregation model, Neural Network is the most common model. Bin Khamis in \cite{BinKhamis_2014} compared the performance of the Neural Network against the Multiple-Linear Regression (MLR). NN achieved a higher $R^2$ value and a lower $MSE$ than the MLR. Comparing the results of the Hedonic model versus the neural network model, the neural network outperforms the Hedonic model by achieving a higher $R^2$ value by 45.348\% and a lower $MSE$ by 48.8441\%. The lack of information in the Hedonic model may be the cause of the poor performance. However, there are some limitations in the Neural Network Model, as the estimated price is not the actual price but it is close to the real one. This is because of the difficulty in obtaining the real data from the market. Also, the time effect plays an important role in the estimation process which Neural Networks cannot handle automatically. This implies that the property price is affected by many other economic factors that are hard to be included in the estimation process.
In this paper, we want to investigate the impact of aggregating visual features with textual attributes on the estimation process. Two estimation models will be examined: the SVR and the NN.

%\vspace{-1em}
\section{\uppercase{Dataset description}}
\vspace{-1em}

\begin{figure}[h!]
\begin{center}
\includegraphics[width=1\columnwidth]{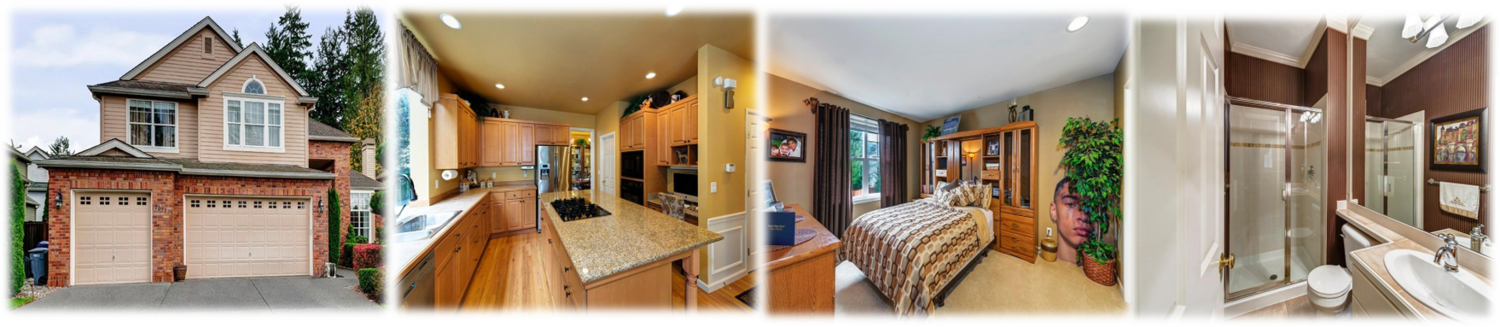}
\caption{\label{fig:figure1} Sample house from \cite{realtor}, where it is represented by 4 images for the frontal side, the kitchen, the bedroom and the bathroom.
}
\end{center}
\end{figure}

The collected dataset is composed of 535 sample houses from California State in the United State. Each house is represented by both visual and textual data. The visual data is a set of 4 images for the frontal image of the house, the bedroom, the kitchen and the bathroom as shown in figure \ref{fig:figure1}. The textual data represent the physical attributes of the house such as the number of bedrooms, the number of bathrooms, the area of the house and the zip code for the place where the house is located. This dataset was collected and annotated manually from publicly available information on websites that sell houses. There are no repeated data nor missing ones. The house price in the dataset ranges from \$22,000 to \$5,858,000. Table 1 contains some statistical details about our dataset. This dataset is publicly available for further research on \cite{ourDataset}.

 %%%%%%%%%%%%%%%%%%%%%%%%%%%%%%%%%
\begin{table}[h]
\caption{Some details about our benchmark houses dataset.}\label{tab:example1} \centering
\begin{adjustbox}{width=0.485\textwidth}
\begin{tabular}{|c|c|c|c|c|}
  \hline
		Detail & Average & Minimum & Maximum \\
		\hline
		\shortstack{House price \\(USD)} & \$589,360 & \$22,000 & \$5,858,000 \\
		\hline
		\shortstack{House area \\(sq. ft.)}& 2364.9  & 701  & 9583  \\
		\hline
		\shortstack{Number of \\bedrooms} & 3.38 & 1 & 10 \\
		\hline
		\shortstack{Number of \\bathrooms} & 2.67 & 1 & 7 \\
        		\hline
       		 \shortstack{Images \\resolution} & 801x560 & 250x187 & 1484x1484 \\
\hline

\end{tabular}
\end{adjustbox}
\end{table}

\section{{\uppercase{Proposed baseline system}}}

\begin{figure}[h!]
\begin{center}
\includegraphics[width=1\columnwidth]{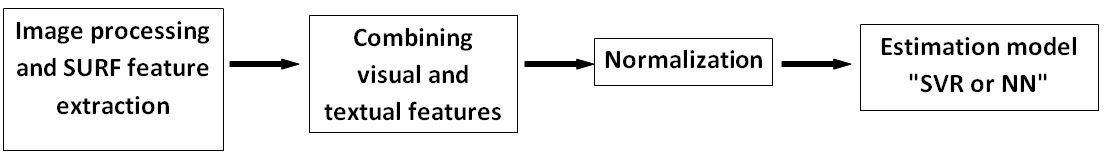}
\caption{\label{fig:figure2} Proposed system processing pipeline.%
}
\end{center}
\end{figure}

\begin{figure}[h!]
\begin{center}
\includegraphics[width=1\columnwidth]{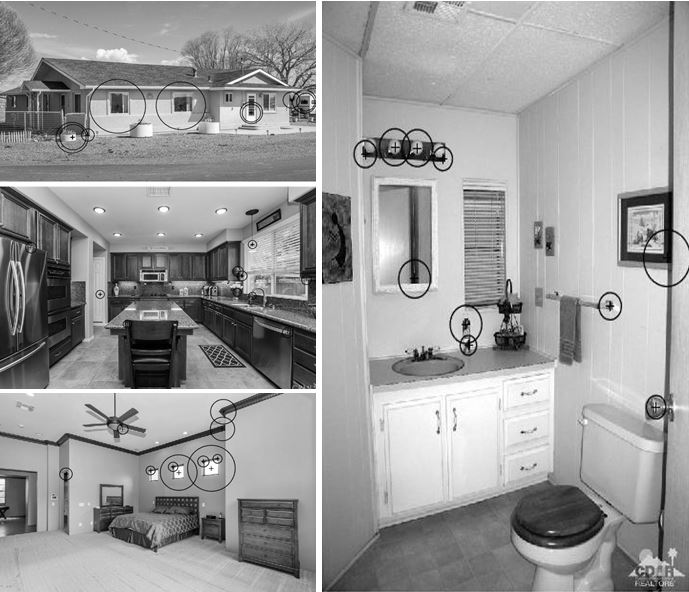}
\caption{\label{fig:figure9} Example for the extracted SURF features from the dataset.%
}
\end{center}
\end{figure}

The main aim of our research is to test the impact of including visual features of houses to be used for the houses' prices estimation. Also, we tried to find the relationship between the number visual features and the accuracy of the estimation using Support Vector Regression and Neural Networks Model.
As shown in figure \ref{fig:figure2}, our system has different processing stages, each of them is represented by a module block. The first module in our system is image processing where the histogram equalization technique \cite{Kapoor2015} is used to increase the global contrast of the dataset images. This technique resulted in better distribution of the color intensity among all the images and allowed the areas of lower local contrast to gain high contrast by effectively spreading out the most frequent intensity values. After that, the Speeded Up Robust Features (SURF) extractor \cite{bay2008speeded} is used for to extract the visual features from the images. SURF uses an integer approximation of the determinant of Hessian blob detector, which can be computed with 3 integer operations using a precomputed integral image. Its feature descriptor is based on the sum of the Haar wavelet response around the point of interest. These can also be computed with the aid of the integral image \cite{bay2008speeded}. In this step, the strongest $n$ features were extracted from each image of the house, then these features were concatenated together in a vector format along with the textual attributes of the same house in a specific order to represent the total features of this house. Figure  \ref{fig:figure9} is an example for the extracted SURF features from the 4 images of a house in the dataset. The extracted features emphasize corners, sharp transitions and edges. It was found visually that these features mark interest points in the images as shown in the frontal image of the house, where the windows were selected as important features. The value for the extracted features $n$ varied from one experiment to another as will be explained in section 5. SURF feature extractor produced better results compared to the Scale Invariant Feature Transform (SIFT) extractor \cite{lowe2004distinctive} and it was also faster therefore, it was used in all of our experiments. In the last module, the aggregated features are passed to one of the estimation modules: either the SVR or the NN after normalization. Normalization is a preprocessing technique where data is scaled between the range of 0 and 1. The formula used for normalization is:

\begin{eqnarray}
\textrm{$z_{i}$} &=\frac{x_{i}-min(x)}{max(x)-min(x)}
\end{eqnarray}
Where $ x = (x_{1},...,x_{n})$ and $z_{i}$ is the $i^{th}$ normalized data point.

Each estimation model has its own architecture and parameters.

\subsection{Support Vector Regression (SVR)}

Support Vector Machines are Machine Learning approaches for solving multi-dimensional function estimation and regression problems. SVMs are derived from the statistical learning theory and they are based on the principle of optimal separation of classes. SVMs use high dimensional feature space to learn and yield prediction functions that are expanded on a subset of support vectors \cite{basak2007support}.
There are two main categories for the SVMs: Support Vector Classification (SVC) and Support Vector Regression (SVR). In SVCs, the SVMs try to separate the classes with the minimum generalization error if the classes are separable. If the classes are not seperable, SVMs try to get the hyperplane that maximizes the margin and reduces the misclassification error. In SVRs, Vapnik in \cite{sain1996nature} introduced an alternative intensive loss function $\epsilon$ that allows the margin to be used for regression. The main goal of the SVR is to find a function $f(x)$ that has at most $\epsilon$ deviation from the actually obtained targets for all the training data and at the same time as flat as possible. In other words, the error of the training data has to be less than $\epsilon$ that is why the SVR depends only on a subset of the training data because the cost function ignores any training data that is close or within $\epsilon$ to the model prediction \cite{svrKernels,basak2007support}.
A deep explanation of the underlying mathematics of the SVR is given in \cite{basak2007support}. It also points out that the SVR requires a careful selection of the kernel function type and the regularization parameter (C). The kernel function can efficiently perform non-linear regression by implicitly mapping the inputs into a higher dimensional feature space to make it possible to perform linear regression. The (C) parameter determines the trade-off between the flatness of the function and the amount by which the deviations to the error more than \textbf{\textit{$\epsilon$}} can be tolerated \cite{svrKernels}. In our experiments, the \textit{Histogram Intersection Kernel} was chosen as the kernel type and the optimal value for the parameter (C) was obtained after several experiments on the dataset to obtain the best result. Histogram Intersection is a technique proposed in \cite{Swain1991} for color indexing with application to object recognition and it was proven in \cite{1247294} that it can be used as a kernel for the SVM as an effective representation of color-based recognition systems that are stable to occlusion and to change of view.

The metrics for evaluating the performance of the SVR are the \textit{coefficient of determination} ($R^2$) and \textit{the Mean Squared Error} ($MSE$).

\subsection{Neural Networks (NNs)}

\begin{figure}[h!]
\begin{center}
\includegraphics[width=1\columnwidth]{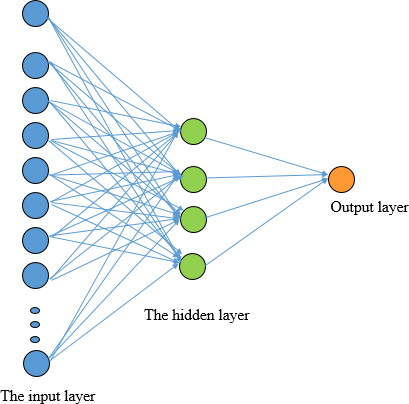}
\caption{\label{fig:figure3} General structure of the Neural Network where it consists of 3 layers: the input layer, one hidden layer of 4 neurons and the output layer.%
}
\end{center}
\end{figure}

Neural Networks are artificial intelligence models that are designed to replicate the human brain. NNs typically consist of layers as shown in figure \ref{fig:figure3} .These layers are formed by interconnected processing units called neurons where the input information is processed. Each neuron in a layer is connected to the neurons in the next layer via a weighted connection. This weighted connection $W_{ij}$ is an indication of the strength between node $i$ where it is coming from and node $j$ where it is going. A three layer NN is shown in figure \ref{fig:figure3}. The structure of the NN is an input layer, one or more hidden layers and an output layer. Hidden layers can be called as feature detectors because the activity pattern in the hidden layer is an encoding of what the network thinks are the significant features of the inputs. When combining the hidden layers features together, the output unit can perform more complex classification/regression tasks and solve non-linear problems. NNs that have one or more hidden layers are used for solving non-linear problems. The architecture of the NN depends on the complexity of the problem.

Each neuron performs a dot product between the inputs and the weights and then it applies an activation function. There are many different types of activation functions. The most common activation function that is also used in our experiments is the sigmoid activation function $f\left(x\right)=\frac{1}{1+e^{-x}}$. The advantage of this function is that it is  easy to differentiate which dramatically reduces the computation burden in the training. Both the inputs and the outputs of the sigmoid function are in the range between 0 and 1 that is why we had to normalize the data before starting the NN.
Our NN was trained using Levenberg--Marquardt algorithm (LMA) \cite{gavin2011levenberg} which is a technique used to solve non-linear least square problems. The Levenberg-Marquardt method is a combination of two minimization methods: the gradient descent method and the Gauss-Newton method. In the gradient descent method, the sum of the squared errors is reduced by updating the parameters in the steepest-descent direction. In the Gauss-Newton method, the sum of the squared errors is reduced by assuming the least squares function is locally quadratic, and finding the minimum of the quadratic. The Levenberg-Marquardt method acts more like a gradient-descent method when the parameters are far from their optimal value, and acts more like the Gauss-Newton method when the parameters are close to their optimal value.
We used \textit{coefficient of determination} ($R^2$) and \textit{the Mean Squared Error} ($MSE$) for evaluating the performance of the NN on our dataset and to compare the results with the \cite{Lichman:2013} housing dataset

\subsection{Performance evaluation}

\subsubsection{Mean Square Error}

Mean Square Error is a measure for how close the estimation is relative to the actual data. It measures the average of the square of the errors deviation of the estimated values with respect to the actual values. It is measured by:
\begin{eqnarray}
\textrm{MSE} &=\frac{1}{n} \sum\limits_{i=1}^n(\hat y-y)^2
\end{eqnarray}
where $\hat y$ is the estimated value from the regression and $y$ is the actual value. The lower the MSE, the better the estimation model.

\subsubsection{The coefficient of determination $R^2$}

The coefficient of determination is a measure of the closeness of the predicted model relative to the actual model. It is calculated a set of various errors:
\begin{eqnarray}
\textrm{SSE} &=\sum\limits_{i=1}^n(\hat y_{i}-y_{i})^2\\
\textrm{SST} &=\sum\limits_{i=1}^n(\bar y-y_{i})^2
\end{eqnarray}
$SSE$ is the Sum of Squares of Error and $SST$ is the Sum of Squares Total. The R-squared value is calculated by:
\begin{eqnarray}
R^2 = 1 - \frac{SSE}{SST}
\end{eqnarray}
The value of $R^2$ ranges between 0 and 1, the higher the value, the more accurate the estimation model.

\section{{\uppercase{Experiments and results}}}

In this section, we will describe the experiments we have done in both estimation models: SVR and NN and compare the NN with the \cite{Lichman:2013} Housing dataset.
\subsection{SVR experiments}
In the SVR model, 428 houses were used for training which is 80\% of the dataset and 107 houses were used for testing which is 20\% of the dataset. The SVR was trained and tested on different number of the extracted SURF features each time to find the relationship between the number of features and the accuracy of the estimation model. 16 different cases were tested starting with training and testing with the textual attributes only with no visual features and moving forward to extracting more SURF features up to 15. In our experiments, the Histogram Intersection Kernel was chosen as the kernel type and the optimal value for the parameter (C) was obtained after several experiments on the dataset to obtain the best result.
Figures \ref{fig:figure5} and \ref{fig:figure6} in section 5.2 show that performance of the SVR increases with adding more visual features till it reaches 9 visual features where the model achieves the lowest $MSE$ value of 0.0037665 and the highest $R-Value$ of 0.78602. Then, the SVR performance started to deteriorate gradually after reaching its highest point at 9 features.

\subsection{Neural Networks experiments}

\begin{figure}[h!]
\begin{center}
\includegraphics[width=1\columnwidth]{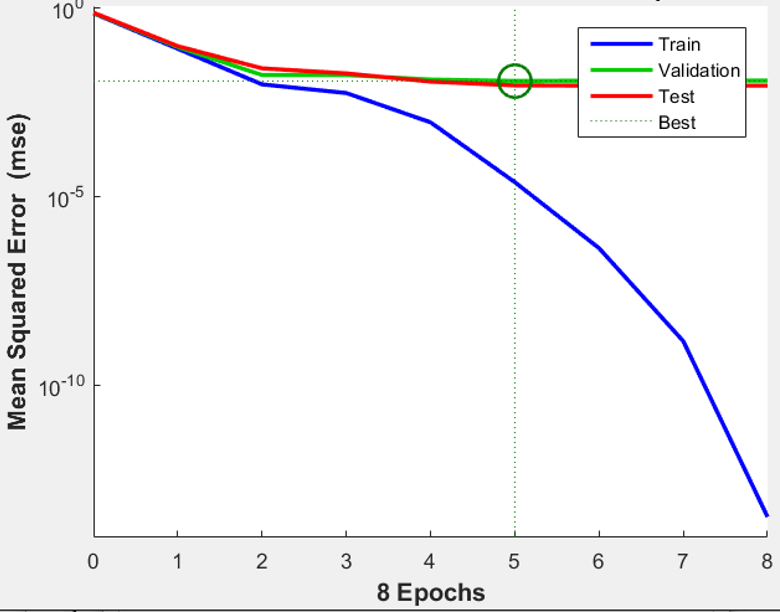}
\caption{\label{fig:figure4} Performance graph shows the $MSE$ of the network per epoch on training, validation and testing sets.%
}
\end{center}
\end{figure}

As shown in figure \ref{fig:figure3}, we adopted a fully connected architecture with one 4-units hidden layer. The problem was expected to be non-linear that is why the networks has hidden layers. The number of hidden nodes was chosen to be somewhere between the number of input nodes and output nodes and by trying different number of neurons in the hidden layer, it was proven that having 4 neurons is the optimal architecture. Our neurons had sigmoid activation function and trained with the Levenberg Marquardt variation of the error-back-propagation technique. This architecture produced the best results during our experiments. We divided our dataset into three pieces: 70\% for training, 15\% for validation, and 15\% for testing.
To avoid over-fitting, we have stopped the training after 5 epochs, a measure of the number of times all of the training vectors are used once to update the weights, because the validation error started to increase. Figure \ref{fig:figure4} shows the performance of the Network highlighting the training, validation and test MSEs and when the training process was stopped to avoid over-fitting.

\begin{figure}[h!]
\begin{center}
\includegraphics[width=1\columnwidth]{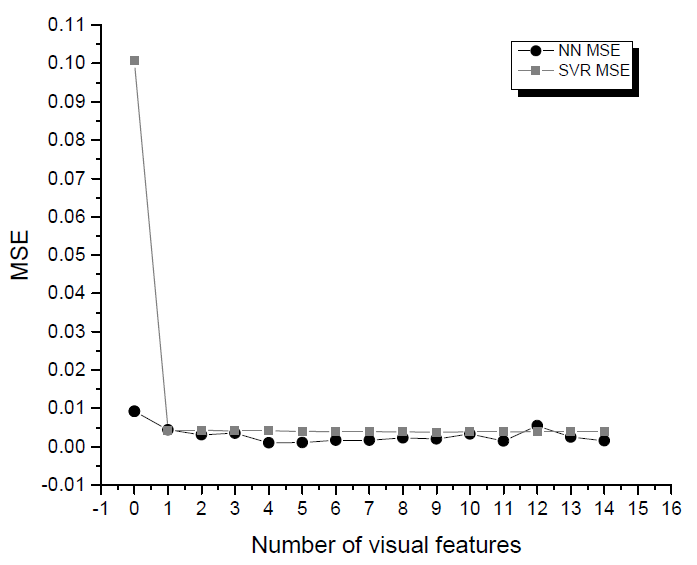}
\caption{\label{fig:figure5} The relationship between the number of features and the MSE in the NN model and the SVR model.%
}
\end{center}
\end{figure}

\begin{figure}[h!]
\begin{center}
\includegraphics[width=1.1\columnwidth]{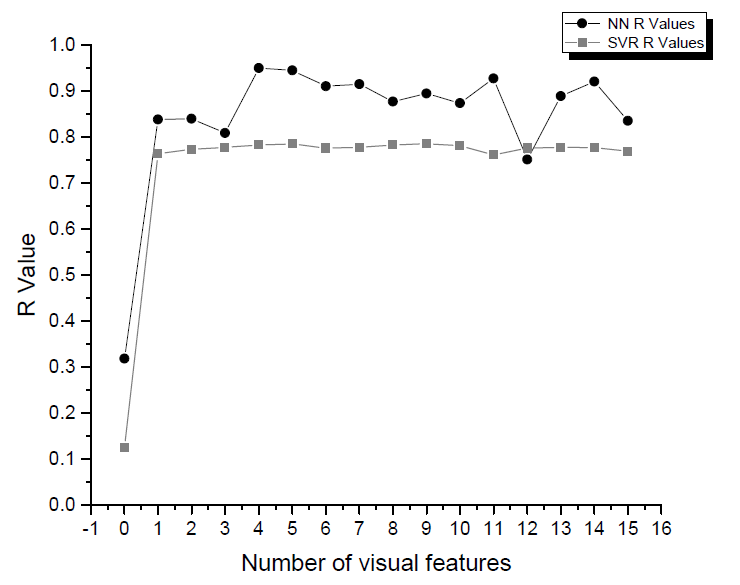}
\caption{\label{fig:figure6} The relationship between the number of features and the R-Value in the NN model and the SVR model.%
}
\end{center}
\end{figure}

\begin{figure}[h!]
\begin{center}
\includegraphics[width=0.9\columnwidth]{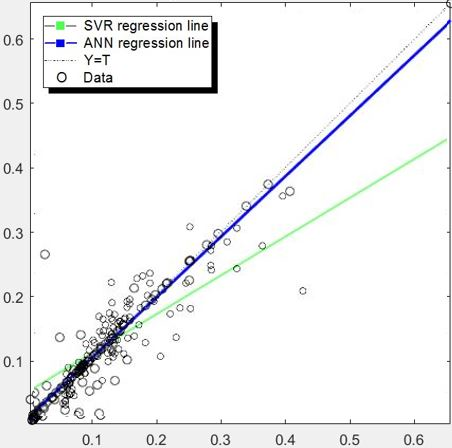}
\caption{\label{fig:figure8} The regression line for the SVR and the NN.%
}
\end{center}
\end{figure}

Figures \ref{fig:figure5} and \ref{fig:figure6} show that combining 4 SURF features with the textual attributes results in achieving the highest $R-Value$ of 0.95053 and the least $MSE$ of 0.000959. In the NN model, the $MSE$ starts very high with no visual features and with increasing the visual features, the $MSE$ starts to decrease till it reaches its minimum value at 4 features, and then it gradually starts to increase till 16. Figure \ref{fig:figure5} shows that the NN outperforms the SVR model by achieving a lower $MSE$ by 76.02\%. Also, figure \ref{fig:figure6} shows that the NN achieved a higher $R-Value$ by 21.05\% than the SVR. Also figure \ref{fig:figure8} shows that the regression line produced by the NN is more accurate because the estimated values are much closer to the actual data.

\subsection{Our NN model on \cite{Lichman:2013} Housing dataset}

To rule out data dependency, we have tested our model on the \cite{Lichman:2013} benchmark housing dataset that has 506 houses, each with 13 textual attributes such as average number of rooms per dwelling, age of the house, full-value property tax, etc. We compared our results with \cite{BinKhamis_2014} model that used NN to estimate the house price based on textual features only. We replicated their model to be able to compare the results of both training and testing instead of the training only which was reported in the paper. We compared the $MSE$ and $R-Value$ in both training and testing and our model outperforms their model. Average prices were used while calculating the $MSE$ to compare the results with \cite{Lichman:2013} model that is why the $MSE$ values are larger than the values reported on our dataset. 

The results tabulated in table 2 show that our model achieves an $MSE$ of $9.708 \times 10^6 $ and $R-Value$ of $0.9348$ on the testing set which is better that Bin Khamis's model that achieves an $MSE$ of $1.713 \times 10^9 $ and R-Value of $0.87392$. Our model achieves a lower $MSE$ on the training set by $99.54\%$ and on the testing set by $99.43\%$. It also achieves a higher $R-Value$ by $6.8\%$ on the training set and on the testing set $6.97\%$. These results show that our Neural Network model does not depend on our own dataset.

%%%%%%%%%%%%%%%%%%%%%%%%%%%%%%%%%%%%%%%%%%
\begin{table}[h]
\caption{Comparison between our NN performance and Bin Khamis's model.}\label{tab:example1} \centering
\begin{adjustbox}{width=0.5 \textwidth}
\begin{tabular}{|c|c|c|c|c|}
  \hline
            & \shortstack{Training \\ MSE}                      & \shortstack{Training \\ R-Value}                & \shortstack{\\Testing \\ MSE  }                  & \shortstack{Testing \\ R-Value} \\ \hline
\shortstack{\\Bin Khamis's \\ model} &1.293 E9  & 0.9039  & 1.713 E9  & 0.87392      \\ \hline
\shortstack{\\Our\\ model}   &5.9223 E6 & 0.96537 & 9.708 E6 &  0.9348       \\ \hline
\end{tabular}
\end{adjustbox}
\end{table}

\section{Conclusion}
This paper announces the first dataset, to our knowledge, that combines both visual and textual features for house price estimation. Other researchers are invited to use the new dataset as well. Through experiments, it was shown that aggregating both visual and textual information yielded better estimation accuracy compared to textual features alone. Moreover, better results were achieved using NN over SVM given the same dataset. Additionally, we demonstrated empirically that the house price estimation accuracy is directly proportional with the number of visual features up to some level, where it barely saturated. We believe this optimal number of features depends on the images content. We are currently studying the relationship of the image content to the optimal number of features. In the near future, we are planning to apply deeper neural networks to extract its own features as well as trying other visual feature descriptors, e.g., Local Binary Patterns (LBP).

\vfill
\bibliographystyle{apalike}
{\small
\bibliography{Houses}}

\vfill
\end{document}